\documentclass{article}
\usepackage{conference,times}
\preprintfinalcopy
\usepackage[T1]{fontenc}
\usepackage[utf8]{inputenc}
\usepackage{amsmath,amssymb,amsfonts}
\usepackage{graphicx,booktabs,multirow,array,tabularx}
\usepackage[table]{xcolor}
\usepackage{url}
\usepackage[hidelinks]{hyperref}
\usepackage{placeins}
\usepackage{enumitem}

\graphicspath{{figures/}{./}}
\definecolor{ink}{HTML}{20394E}
\definecolor{fscolor}{HTML}{087F72}
\definecolor{fspale}{HTML}{E7F4EE}
\definecolor{bluepale}{HTML}{EAF1F8}
\definecolor{amberpale}{HTML}{FFF1DF}
\definecolor{papergray}{HTML}{F5F7FA}
\definecolor{draftred}{HTML}{B52D3B}
\newcommand{\method}{\textit{FinishStrong}}
\newcommand{\sg}{\operatorname{sg}}
\newcommand{\IoU}{\operatorname{IoU}}
\newcommand{\Rec}{\operatorname{cIoU}}
\newcommand{\Acc}{\operatorname{Acc}}
\newcommand{\E}{\mathbb{E}}
\newcommand{\ind}{\mathbf{1}}
\newcommand{\Reg}{\operatorname{Reg}}
\newcommand{\best}[1]{\textbf{#1}}

\title{FinishStrong: Learning to End Well after\\Imperfect Start in Video Temporal Grounding}

\author{
  \textbf{Yuda Zou$^{1,2}$, Boxiang Zhou$^2$, Xin Zhou$^2$, Yibo Chen$^2$,} \\
  \textbf{Dejia Song$^2$, Xu Tang$^2$, Yao Hu$^2$, Yongchao Xu$^1$\thanks{Corresponding author.}} \\
  \\
  $^1$School of Computer Science, Wuhan University \\
  $^2$Xiaohongshu Inc. \\
  \texttt{zouyuda@whu.edu.cn}
}

\begin{document}
\maketitle

\begin{abstract}
A poor start lowers the ceiling, but does not negate the value of finishing well. 
In autoregressive video temporal grounding, models systematically exhibit an endpoint performance asymmetry, predicting ends significantly worse than starts. A committed flawed start acts as an immutable physical constraint, irreversibly capping the attainable overlap as a sunk loss. While models should minimize the remaining conditional regret, standard post-training paradigms fail to teach this adaptation. Specifically, teacher-forced SFT suffers from strict prefix exposure bias, and holistic trajectory evaluation in standard GRPO inadvertently penalizes conditionally optimal ends if they follow a poor start. To address this, we introduce \method, a novel two-stage post-training framework explicitly designed to teach optimal continuations. First, Online-start SFT exposes the end prediction to the model's own generated contexts. Second, Conditional-tree GRPO restructures the sampling topology, centering end rewards exclusively within the same-start branch to mathematically isolate the conditional end quality from the inherited start penalty. \method-9B achieves a state-of-the-art 61.7\% overall IoU on TimeLens-Bench, remarkably outperforming flagship closed-source models like Gemini-3.8-Flash and Qwen3.8-Max. Crucially, when forced to continue from imperfect starts, it reduces the subsequent end prediction error by 36.5\%, proving its exceptional ability to end well after flawed starts. Code will be open.
\end{abstract}

\section{Introduction}
\label{sec:intro}
\begin{figure}[t]
\centering
\includegraphics[width=0.9\linewidth]{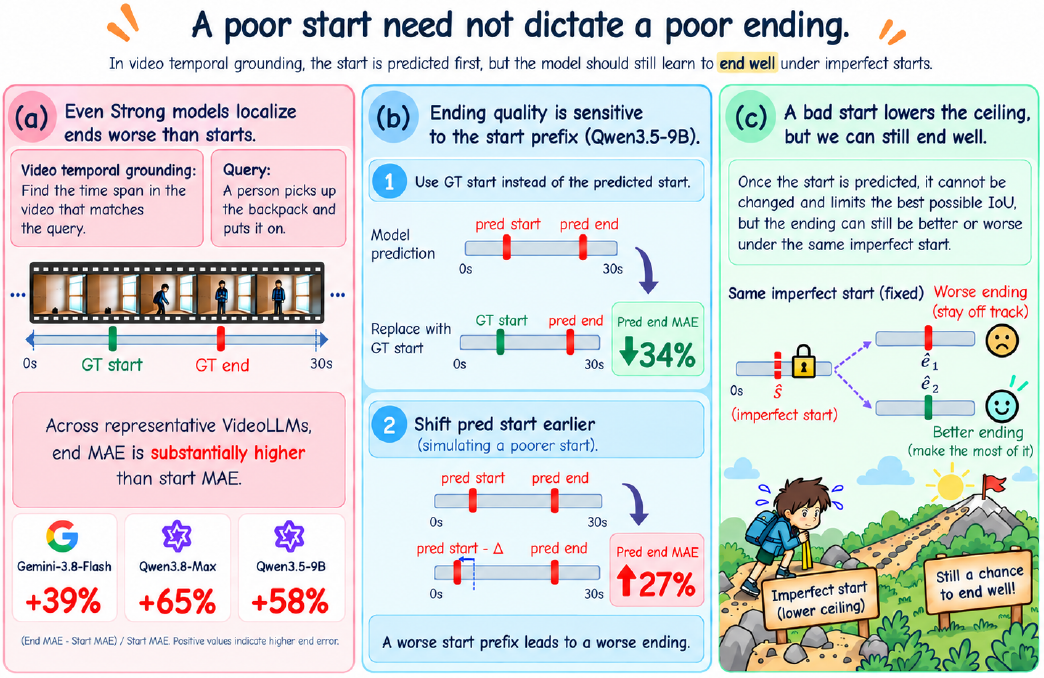}
\caption{\textbf{
(a) Even strong models localize ends worse than starts.
(b) Ending quality is sensitive to the start prefix.
(c) A bad start lowers the ceiling, but we can still end well.}}
\label{fig:diagnosis}
\end{figure}

Video temporal grounding, the fundamental cornerstone for fine-grained video understanding tasks, requires accurately localizing the start and end timestamps of a text-queried event~\citep{qvhighlights, charades-sta, caba2015activitynet,yang2025timeexpert}.
As Vision-Language Models (VLMs) grow more capable, they increasingly tackle this task through an autoregressive paradigm, sequentially generating time boundaries as text tokens~\citep{Zhang_2026_CVPR_timelens,wang2025timer1,li2026videoopd,du2026mllms}.
However, these models consistently exhibit an endpoint performance asymmetry: they are significantly worse at predicting ends than starts. Even in flagship models (e.g., Gemini-3.8-Flash, Qwen3.8-Max), the end mean absolute error (MAE) remains 39\% to 65\% higher than the start MAE (Figure~\ref{fig:diagnosis}a).
This substantial gap is a robust empirical observation across diverse architectures, raising the critical question of what drives this systematic degradation at the endpoint.

To explore the root cause, we experimentally isolate the effect of the generated start. Keeping the model, video, and query fixed, replacing the Qwen3.5-9B model's generated start with the ground-truth (GT) start reduces its end MAE by 34\% (from 11.18 to 7.38 seconds). 
Conversely, perturbing its own predicted start causes the subsequent end MAE to surge by 27\% (Figure~\ref{fig:diagnosis}b). 
These bidirectional interventions prove that a massive portion of end error is directly triggered by the prefix start. 
In autoregressive generation, a committed start acts as an immutable physical constraint. Even following an inaccurate start, the mathematically optimal strategy to maximize remaining temporal overlap (IoU) is still to predict the exact GT end timestamp. 
\textbf{The start determines the ceiling; the end determines how much of it is realized.} 
Conceptually, the model should treat the start error as a sunk loss and focus on precise end localization to minimize conditional regret. 
Instead, models are easily misled by the flawed prefix contexts, causing the severe end prediction errors observed above rather than making the optimal remaining choice.

We trace this failure back to current post-training paradigms, which predominantly teach models how to finish well under ideal conditions, but neglect to train them on adverse ones. Specifically, teacher-forced SFT supplies perfect prefixes, leaving the model unexposed to its own generated errors \citep{bengio2015scheduled,ross2011dagger}. 
Standard GRPO effectively optimizes overall generation by rewarding a good end after a good start, and penalizing a poor end after a poor start. However, its holistic trajectory evaluation inadvertently neglects to reward a conditionally good end following a poor start \citep{shao2024deepseekmath,Zhang_2026_CVPR_timelens}.
Finally, Video-OPD \citep{li2026videoopd} relies on teacher models that suffer from the same endpoint limitations (Figure~\ref{fig:diagnosis}a). Consequently, without explicit training on how to adapt, a flawed start almost guarantees a flawed end.

To address this, we introduce a two-stage post-training framework \method, which fundamentally changes both \emph{where the end learns} and \emph{what it is compared with}. 
First, Online-start SFT exposes the end prediction to the model's own detached starts. This forces the model to learn under self-generated contexts while strictly preserving ground-truth start supervision.
Second, Conditional-tree GRPO restructures the sampling topology without increasing the overall generation budget. By assigning starts their expected downstream value and centering end rewards exclusively within the same-start branch, this design mathematically isolates the conditional end quality from the inherited start penalty.
\method-9B achieves a state-of-the-art 61.7\% overall IoU on TimeLens-Bench. Furthermore, under rigorous shared-prefix stress tests, it demonstrates exceptional robustness to imperfect starts, reducing the global conditional end MAE by 36.5\%. Our main contributions are:
\begin{enumerate}[label=\arabic*., leftmargin=*, noitemsep, topsep=0pt]
    \item To the best of our knowledge, we are the first to expose the systematic endpoint asymmetry in autoregressive video temporal grounding, revealing how standard post-training paradigms inherently fail to optimize the conditional regret remaining after a flawed start.
    \item We propose \method, a novel post-training framework to teach models how to make optimal continuations after imperfect starts. \method-9B establishes a new state-of-the-art, notably outperforming flagship closed-source models like Gemini-3.8-Flash and Qwen3.8-Max.
\end{enumerate}

\section{Related work}
\label{sec:related}

\textbf{Video temporal grounding.}
Video temporal grounding~\citep{cao2026one,zou2026decoupling} has been approached through dedicated localization architectures and generative video--language models.
BAM-DETR adopts a boundary-oriented formulation for moment localization~\citep{lee2024bam}, while TimeSuite, TRACE, and Grounded-VideoLLM improve temporal modeling and grounded generation through specialized tuning and representations~\citep{zeng2024timesuite,guo2024trace,wang2024grounded_videollm}.
Recent post-training work includes reinforcement learning in Time-R1~\citep{wang2025timer1} and systematic investigations of data quality, timestamp encoding, and training recipes in TimeLens~\citep{Zhang_2026_CVPR_timelens,zhu2026timelens2}.
Our work addresses a complementary question within autoregressive grounding: how to learn an accurate end prediction after the model has already committed to a possibly inaccurate start.

\textbf{Learning under model-generated prefixes.}
Scheduled sampling introduces model-generated tokens during supervised sequence training~\citep{bengio2015scheduled}, while DAgger obtains expert supervision on states visited by the learner~\citep{ross2011dagger}.
Both motivate reducing the mismatch between training contexts and those encountered at inference.
For temporal grounding, Video-OPD provides dense teacher feedback on student-generated trajectories through on-policy distillation~\citep{li2026videoopd}.
Online-start SFT shares the principle of learning under generated contexts but exploits a task-specific property: for a valid fixed start $p<e^\star$, the annotated end $e^\star$ maximizes temporal IoU (Section~\ref{sec:diagnosis}).
We therefore supervise the end directly under the model's detached start, without an external teacher, while retaining ground-truth start supervision.

\textbf{Decision-conditioned credit assignment.}
GRPO evaluates sampled completions relative to a prompt-level reward group~\citep{shao2024deepseekmath}.
Conditioning the baseline on the current decision state is a standard principle in policy-gradient methods~\citep{schulman2015gae}.
Tree-structured optimization has also been explored: VideoMiner's T-GRPO learns hierarchical key-frame exploration~\citep{cao2025videominer}, and TreeGRPO propagates prompt-normalized leaf advantages through denoising trees to obtain per-step credit~\citep{ding2026treegrpo}.
Conditional-tree GRPO instead separates between-start and within-start comparisons: starts are evaluated by mean downstream IoU, whereas end rewards are centered only among continuations sharing the same start.
This within-start contrast compares relative conditional end regret rather than outcomes with different attainable ceilings.
We retain the original IoU reward without ceiling normalization and use the tree only during training; inference remains one greedy start followed by one greedy end.

\section{The imperfect-start problem}
\label{sec:diagnosis}

\subsection{Sunk loss and the conditional overlap ceiling}
For a video--query input $x$, let the ground-truth (GT) interval be $[s^\star,e^\star]$ with length $L^\star>0$. The autoregressive policy factorizes sequentially:
\begin{equation}
\pi_\theta(s,e\mid x)=\pi_\theta(s\mid x)\,\pi_\theta(e\mid x,s).
\label{eq:factor}
\end{equation}
A committed start $p$ acts as an immutable constraint. The maximum achievable overlap (IoU) for any subsequent end is bounded by a conditional ceiling:
\begin{equation}
C_{\max}(p)=\max_{e'>p}\IoU([p,e'],[s^\star,e^\star])=
\begin{cases}
L^\star/(e^\star-p),&p<s^\star,\\
(e^\star-p)/L^\star,&s^\star\le p<e^\star.
\end{cases}
\label{eq:ceiling}
\end{equation}
This ceiling is uniquely attained at $e^\star$. Crucially, a flawed start permanently lowers it: for a GT event $[20,40]$, a late start $p=32$ caps the achievable IoU at $0.4$ (attained only at $e=40$). The model must optimally utilize this remaining opportunity, despite the irreversible start error.

Let $R(p,e)=\IoU([p,e],[s^\star,e^\star])$. The shortfall from a perfect interval ($R=1$) decomposes orthogonally:
\begin{equation}
1-R(p,\hat e)
=\underbrace{1-C_{\max}(p)}_{\text{committed start loss}}
+\underbrace{C_{\max}(p)-R(p,\hat e)}_{\text{conditional end regret }\Reg_e(p,\hat e)}.
\label{eq:sunk}
\end{equation}
The first term is an irreversible ``sunk loss''. Only the conditional end regret $\Reg_e$ can be minimized by the end policy $\pi_\theta(e\mid x,p)$. Thus, \textbf{maximizing IoU given an imperfect start is mathematically equivalent to minimizing the conditional end regret.}

\begin{figure}[t]
\centering
\includegraphics[width=0.9\linewidth]{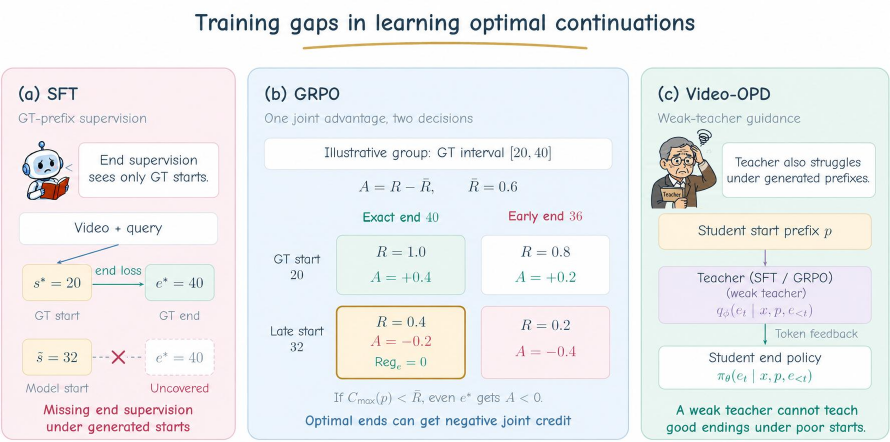}
\caption{\textbf{How standard paradigms interact with optimal continuations.} 
(a) \textbf{SFT} typically trains on perfect starts, leaving generated errors less covered. 
(b) \textbf{GRPO} evaluates trajectories holistically. If a flawed start lowers the ceiling below the group mean ($C_{\max} < \bar{R}$), even an optimal end ($\Reg_e=0$) might still receive a negative advantage. 
(c) \textbf{Video-OPD} provides feedback that is naturally influenced by the teacher's own conditional capabilities.}
\label{fig:fail}
\end{figure}

\subsection{The challenge of teaching optimal continuations}
While standard post-training improves overall generation, explicitly minimizing $\Reg_e$ poses a structural nuance (Figure~\ref{fig:fail}).

\textbf{SFT faces exposure bias.} Because standard SFT typically optimizes ends conditioned on perfect GT starts, the model may have fewer opportunities to learn how to adapt to its own generated errors.

\textbf{GRPO evaluates trajectories holistically.} Standard GRPO assigns a joint scalar advantage $A=R-\bar R$. As mapped in Figure~\ref{fig:fail}b, this effectively guides extreme cases. However, the mixed case presents a subtle challenge. If a flawed start lowers the ceiling below the group mean ($C_{\max}(p) < \bar{R}$), the interval's maximum score is inherently constrained. Consequently, even a zero-regret continuation ($\Reg_e=0$) might still receive a negative advantage. In such specific scenarios, holistic evaluation may inadvertently discourage these branches, making it harder to explicitly reward the effort to finish well.

\textbf{Video-OPD relies on teacher capabilities.} On-policy distillation provides dense feedback \citep{li2026videoopd}. Since the feedback quality depends on the teacher at that start, the student's conditional adaptation is naturally influenced by the teacher's own endpoint asymmetry (Figure~\ref{fig:diagnosis}a).

\section{Method}
\label{sec:method}

\begin{figure}[t]
\centering
\includegraphics[width=0.9 \linewidth]{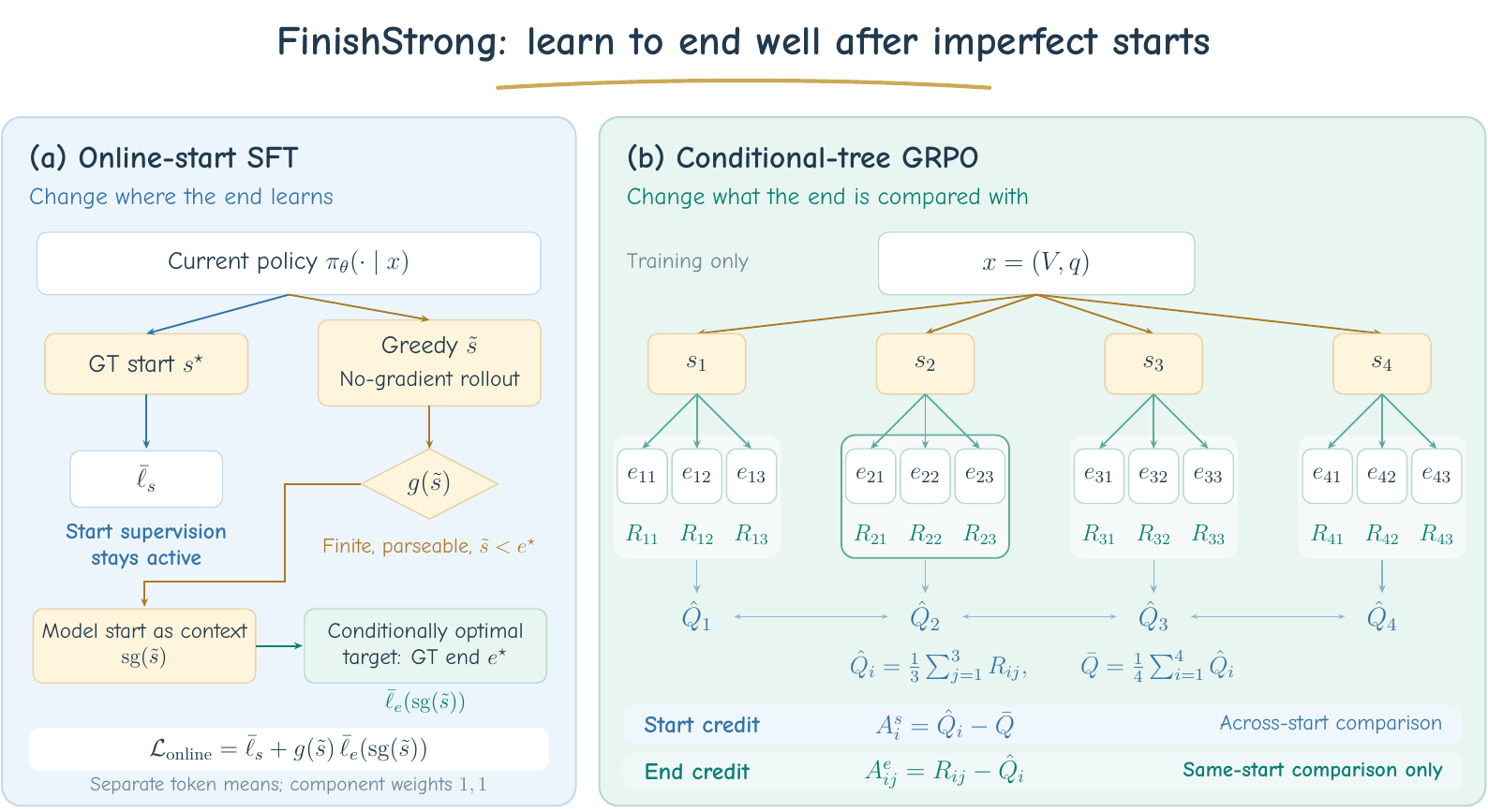}
\caption{\textbf{\method: learn to end well after imperfect starts.} \textbf{(a) Online-start SFT} teaches ends under detached model starts, retaining ground-truth supervision. \textbf{(b) Conditional-tree GRPO} samples 4 starts, each with 3 ends. Start advantages ($A^s$) use average downstream value; end advantages ($A^e$) center exclusively within their branch.}
\label{fig:method}
\end{figure}

To overcome the training gaps identified in Section~\ref{sec:diagnosis}, we must fix both \emph{where} the end learns and \emph{what} it is compared with. Standard teacher-forcing hides generated errors, while joint-credit RL penalizes optimal continuations under poor ceilings. We introduce \method, a two-stage post-training pipeline that systematically resolves these issues while retaining a learning signal for start quality (Figure~\ref{fig:method}). First, Online-start SFT forces the end prediction to learn within the model's self-generated context. Second, Conditional-tree GRPO restructures credit assignment so that end tokens are evaluated strictly against sibling continuations from the identical start.

\subsection{Online-start SFT: Train in the context the model creates}
\label{sec:online}
To break the reliance on perfect prefixes, we expose the end-generation loss to the model's actual sampling distribution. Before each microbatch update, the current policy greedily generates a start $\tilde s$ without tracking gradients. Let $\bar\ell_s$ be the prefix cross-entropy given the GT start, and $\bar\ell_e(p)$ be the conditional end cross-entropy given a rendered start $p$. We optimize:
\begin{equation}
\mathcal L_{\mathrm{online}}=\bar\ell_s+g(\tilde s)\,\bar\ell_e\!\left(\sg(\tilde s)\right),
\label{eq:online}
\end{equation}
where $g(\tilde s)=\ind[\tilde s\text{ finite and parseable},\ \tilde s < e^\star]$.
The stop-gradient ($\sg$) operator strictly limits $\tilde s$ to serving as conditional context, preventing the policy from inadvertently optimizing toward its own start errors. 

Empirically, $g(\tilde s)=1$ holds for over 90\% of generated starts in early updates, ensuring data-efficient conditional training. The true-start supervision ($\bar\ell_s$) explicitly penalizes the remaining invalid starts, resolving them naturally over time. (Crucially, while SFT masks severely late starts ($g=0$) to avoid teaching mathematically impossible continuations, our subsequent RL stage fully samples and heavily penalizes them via low rewards, encouraging a robust global policy.)

\subsection{Conditional-tree GRPO: Align credit with the remaining decision}
\label{sec:tree}
To isolate conditional end quality from the inherited start ceiling, we restructure the reinforcement sampling topology. For each input $x$, we sample $K=4$ starts $s_i\sim\pi_\theta(\cdot\mid x)$ and $M=3$ end continuations $e_{ij}\sim\pi_\theta(\cdot\mid x,s_i)$ per start. \textbf{This $4 \times 3$ structure is designed to match the nominal token generation budget of standard GRPO (group size 8)}, since 4 starts and 12 ends generate a comparable number of tokens as 8 full sequence rollouts ($4 \times 0.5 + 12 \times 0.5 = 8$). Consequently, this topological shift does not significantly increase the overall sampling overhead.

Each of the twelve resulting intervals receives the original task reward. We define the branch mean and the mean across branches as:
\begin{equation}
R_{ij}=R(s_i,e_{ij}),\qquad
\hat Q_i=\frac1M\sum_{j=1}^{M}R_{ij},\qquad
\bar Q=\frac1K\sum_{i=1}^{K}\hat Q_i.
\label{eq:branch}
\end{equation}
A start is evaluated by the \emph{mean} downstream reward of its sampled ends. Using inclusive group-mean advantages, we formulate the start and end credits as:
\begin{equation}
A_i^s = \hat Q_i-\bar Q, \qquad A_{ij}^e = R_{ij}-\hat Q_i.
\label{eq:advantages}
\end{equation}
Start and delimiter tokens receive $A_i^s$, while end-continuation tokens receive $A_{ij}^e$. The comparison sets strictly follow the chronological decisions: start branches compete based on downstream expected quality, and end continuations compete exclusively within their shared-start branch. 

\textbf{Decision-aligned credit via within-start comparisons.}
For a balanced $K\times M$ tree, the joint mean equals $\bar Q$,
yielding the exact decomposition:
\begin{equation}
\underbrace{R_{ij}-\bar Q}_{A_{ij}^{\mathrm{joint}}}
=\underbrace{\hat Q_i-\bar Q}_{\text{between-start credit }A_i^s}
+\underbrace{R_{ij}-\hat Q_i}_{\text{within-start credit }A_{ij}^e}.
\label{eq:creditdecomp}
\end{equation}
Let $\Delta_{ij}=C_{\max}(s_i)-R_{ij}$ and
$\bar\Delta_i=M^{-1}\sum_j\Delta_{ij}$.
Then $A^e_{ij}=\bar\Delta_i-\Delta_{ij}$, so end credit compares
relative conditional regret within the same start.
The start-dependent ceiling cancels in this contrast without
rescaling the original IoU reward.
A best sampled end receives positive credit whenever sibling
rewards are not all equal.
This identity does not by itself guarantee lower gradient variance
or convergence to a conditionally optimal policy.

\textbf{Training changes; inference does not.}
Following standard recipes \citep{Zhang_2026_CVPR_timelens}, Online-start SFT precedes Conditional-tree GRPO. \method\ uses neither an external teacher nor an altered, ceiling-normalized reward function. Crucially, the detached context generation and tree-based credit assignments are applied exclusively during training. At inference, the model retains the standard autoregressive paradigm—generating one greedy start and one greedy end. Thus, \method\ achieves robust conditional adaptation with zero additional computational overhead or search latency during deployment.

\section{Experiments}
\label{sec:experiments}
\subsection{Evaluation design}
\textbf{Datasets and baselines.}
TimeLens-Bench~\cite{Zhang_2026_CVPR_timelens} comprises 3,363 Charades-TimeLens, 4,500 ActivityNet-TimeLens, and 1,541 QVHighlights-TimeLens queries. Our primary global metric, ALL IoU, is the macro-average across all 9,404 queries. We compare against standard post-training recipes (SFT, GRPO) and state-of-the-art temporal grounding frameworks (TimeLens and Video-OPD). We reproduce TimeLens-9B and Video-OPD-9B using their official codebases to ensure strict comparability. A single, fixed \method-9B checkpoint is evaluated across all tests.

\textbf{Inference protocol.}
All models operate under greedy decoding with a visual sampling rate of 2 FPS, processing up to 448 frames (a maximum visual budget of 14,336 tokens). Standard temporal grounding evaluates unconditional performance using Recall@1 (at IoU thresholds 0.3, 0.5, and 0.7) and mean IoU. Conversely, in \emph{shared-prefix tests}, we forcefully supply an identical, mathematically defined start and delimiter to all models, compelling them to generate only the end token.

\begin{table}[t]
\caption{\textbf{Comparison with proprietary and open-source models on temporal video grounding benchmarks, TimeLens-Bench~\cite{Zhang_2026_CVPR_timelens} (\%).} R@$\tau$ is Recall@1 at IoU threshold $\tau$; ALL is sample-averaged IoU. \textbf{Bold} marks the best results.}
\label{tab:sota}
\centering\scriptsize
\setlength{\tabcolsep}{2.1pt}
\resizebox{\linewidth}{!}{%
\begin{tabular}{@{}lccccccccccccc@{}}
\toprule
\rowcolor{bluepale}
& \multicolumn{4}{c}{Charades-TimeLens} & \multicolumn{4}{c}{ActivityNet-TimeLens} & \multicolumn{4}{c}{QVHighlights-TimeLens} & ALL\\
\cmidrule(lr){2-5}\cmidrule(lr){6-9}\cmidrule(lr){10-13}\cmidrule(l){14-14}
Model & R@.3 & R@.5 & R@.7 & IoU & R@.3 & R@.5 & R@.7 & IoU & R@.3 & R@.5 & R@.7 & IoU & IoU\\
\midrule
\multicolumn{14}{l}{\emph{Proprietary Models}}\\
GPT-5 & 59.3 & 42.0 & 22.0 & 40.5 & 57.4 & 44.9 & 30.4 & 42.9 & 72.4 & 60.4 & 46.4 & 56.8 & 44.3\\
Qwen3.5-Plus & 71.4 & 51.1 & 27.9 & 49.7 & 74.2 & 63.2 & 43.8 & 56.4 & 82.0 & 73.8 & 58.5 & 67.3 & 55.8\\
Qwen3.8-Flash & 73.3 & 57.6 & 30.4 & 50.5 & 73.7 & 62.1 & 41.4 & 55.2 & 83.2 & 73.3 & 56.8 & 66.5 & 55.3\\
Qwen3.8-Max & 79.5 & 63.1 & 34.6 & 55.2 & 75.8 & 65.4 & 44.6 & 57.4 & 82.8 & 73.4 & 57.3 & 66.6 & 58.1\\
Gemini-2.5-Flash & 68.7 & 56.1 & 30.6 & 48.6 & 66.8 & 57.5 & 41.3 & 52.5 & 78.2 & 69.4 & 55.0 & 64.3 & 53.0\\
Gemini-2.5-Pro & 74.1 & 61.1 & 34.0 & 52.8 & 72.3 & 64.2 & \textbf{47.1} & 58.1 & \textbf{84.1} & \textbf{75.9} & \textbf{61.1} & \textbf{70.4} & 58.2\\
Gemini-3.1-Pro & 80.6 & 65.9 & 36.7 & 57.1 & 75.1 & 64.1 & 43.0 & 56.9 & 83.5 & 73.7 & 58.6 & 68.2 & 58.9\\
Gemini-3.8-Flash & \textbf{86.5} & \textbf{71.9} & \textbf{40.1} & \textbf{61.1} & \textbf{78.2} & \textbf{66.4} & 46.0 & \textbf{59.7} & 83.8 & 74.6 & 57.6 & 67.9 & \textbf{61.6}\\
\midrule
\multicolumn{14}{l}{\emph{Open-Source Models}}\\
TimeSuite & 56.3 & 35.5 & 18.0 & 38.1 & 27.1 & 17.5 & 8.6 & 19.8 & 27.1 & 16.9 & 9.9 & 21.7 & 26.7\\
Grounded-VideoLLM & 43.3 & 28.7 & 13.5 & 30.0 & 39.2 & 29.6 & 19.5 & 30.0 & 43.7 & 33.8 & 22.5 & 33.4 & 30.6\\
VideoChat-Flash-7B & 60.2 & 37.9 & 17.8 & 39.7 & 35.5 & 21.8 & 10.5 & 24.8 & 45.2 & 30.6 & 16.7 & 32.7 & 31.4\\
VideoChat-R1-7B & 51.9 & 30.8 & 11.7 & 33.7 & 35.0 & 23.9 & 11.3 & 25.0 & 29.3 & 19.1 & 9.4 & 21.5 & 27.5\\
TRACE-uni & 38.2 & 22.9 & 10.4 & 28.1 & 44.3 & 35.1 & 22.6 & 33.6 & 49.9 & 40.0 & 29.2 & 39.8 & 32.6\\
Time-R1-7B & 57.9 & 32.0 & 16.9 & 36.6 & 44.8 & 31.0 & 19.0 & 33.1 & 65.8 & 51.5 & 36.1 & 49.2 & 37.0\\
MiMo-VL-7B & 57.9 & 42.6 & 20.5 & 39.6 & 49.3 & 38.7 & 22.4 & 35.5 & 57.1 & 42.6 & 28.4 & 41.5 & 37.9\\
VideoLLaMA3-7B & 65.4 & 37.9 & 17.5 & 42.0 & 46.8 & 31.0 & 16.6 & 33.0 & 51.7 & 37.1 & 22.1 & 39.0 & 37.2\\
Vidi1.5-9B & 23.2 & 13.7 & 9.0 & 19.0 & 50.5 & 39.0 & 26.5 & 38.2 & 64.4 & 54.6 & 41.6 & 51.8 & 33.6\\
InternVideo3-8B & 58.0 & 42.8 & 19.7 & 38.8 & 50.4 & 39.8 & 23.2 & 37.3 & 64.0 & 53.3 & 38.5 & 50.7 & 40.0\\
Molmo2-O-7B & 59.9 & 46.0 & 24.3 & 42.6 & 56.2 & 45.7 & 30.1 & 42.3 & 73.7 & 62.6 & 48.9 & 58.5 & 45.0\\
Qwen3-VL-8B & 69.2 & 53.4 & 27.5 & 48.3 & 62.1 & 51.2 & 34.4 & 46.8 & 74.2 & 64.6 & 49.3 & 59.4 & 49.4\\
TimeLens-8B & 76.6 & 63.0 & 35.2 & 55.2 & 68.9 & 58.4 & 40.6 & 53.2 & 80.2 & 71.6 & 55.5 & 65.5 & 55.9\\
Video-OPD-8B & 74.7 & 49.2 & 32.8 & 53.0 & 66.4 & 50.4 & 39.3 & 52.1 & 77.4 & 64.9 & 54.4 & 64.7 & 54.5\\
\rowcolor{fspale}
\textbf{FinishStrong-8B} & 78.2 & 64.3 & 36.5 & 56.4 & 71.2 & 62.0 & 43.8 & 55.9 & 80.9 & 73.2 & 59.1 & 68.5 & 58.1\\
Qwen3.5-9B & 71.5 & 48.7 & 24.6 & 47.8 & 68.4 & 56.1 & 37.8 & 51.8 & 77.2 & 67.9 & 52.6 & 63.2 & 52.2\\
SFT-9B & 80.3 & 63.8 & 34.3 & 55.7 & 68.8 & 59.8 & 42.3 & 54.7 & 75.3 & 66.8 & 53.3 & 63.3 & 56.5\\
GRPO-9B & 77.7 & 62.5 & 34.1 & 54.9 & 72.8 & 63.4 & 45.2 & 57.3 & 80.8 & 72.3 & 56.8 & 66.7 & 58.0\\
TimeLens-9B & 78.5 & 63.4 & 37.4 & 56.6 & 70.3 & 59.9 & 43.1 & 55.3 & 80.8 & 70.9 & 57.1 & 66.9 & 57.6\\
Video-OPD-9B & 74.8 & 48.9 & 32.8 & 53.6 & 73.6 & 57.2 & 46.3 & 58.5 & \textbf{83.7} & 70.8 & 60.9 & 70.3 & 58.7\\
\rowcolor{fspale}
\textbf{FinishStrong-9B} & \textbf{81.9} & \textbf{67.6} & \textbf{39.0} & \textbf{58.7} & \textbf{75.5} & \textbf{66.6} & \textbf{49.6} & \textbf{60.8} & 83.5 & \textbf{76.2} & \textbf{62.6} & \textbf{71.0} & \textbf{61.7}\\
\bottomrule
\end{tabular}}
\end{table}

\textbf{Two complementary start-perturbation scans.}
To rigorously evaluate conditional recovery, we design two structured prefix interventions. 
The primary \emph{GT-span scan} shifts the start proportional to the event length $L^\star$, defining $p=s^\star+dL^\star$, across a discrete grid $d \in \{-2, -1, \dots, 0.9\}$ (excluding invalid $p<0$). To assess behavior from the far pre-event history, a complementary \emph{video-span scan} scales negative shifts relative to the pre-event duration $s^\star$, defining $p(r) = s^\star(1+r)$ for $r \le 0$ (using $r \in \{-1, \dots, -0.05\}$ to maintain six early evaluation points) and matching the GT-span logic for $r>0$. 
The positive conditions ($d, r > 0$) naturally coincide with $\mathcal D$ and are deduplicated in joint visualizations. This dual-scan design ensures coverage of both localized event-scale errors and macroscopic video-scale errors. Directional summaries equally weight their six nonzero offsets. Full aggregation details are provided in Appendix~\ref{app:protocol}.

\textbf{Conditional metrics.}
Because an imperfect start irreversibly limits achievable overlap, raw IoU inadvertently penalizes models for inherited errors. To isolate the quality of the remaining decision, we report \emph{ceiling-normalized IoU} (cIoU):
\begin{equation}
\Rec(p,\hat e)=\frac{\IoU([p,\hat e],[s^\star,e^\star])}{C_{\max}(p)}.
\label{eq:ciou}
\end{equation}
Computed per sample before averaging, cIoU measures the fraction of attainable overlap actually realized by the chosen end. As complementary indicators, we retain raw IoU, absolute end MAE (in seconds), end accuracy ($\Acc_\tau$) at varying error tolerances $|\hat e-e^\star|\le\tau L^\star$ for $\tau\in\{0.05, 0.10, 0.20, 0.30\}$.

\subsection{Main results: Unconstrained temporal grounding}
\label{sec:sota}

Table~\ref{tab:sota} reports the unconstrained free-decoding evaluation; baseline results are compiled from their technical reports \citep{ OpenAI2025_GPT5, comanici2025gemini-2-5, li2024videochat_flash, li2025videochat_r1, mimo-vl, Qwen3-VL, qwen2026qwen35} and TimeLens \cite{Zhang_2026_CVPR_timelens}. We also include VideoLLaMA3~\citep{zhang2025videollama3}, Vidi1.5~\citep{vidi2025vidi}, InternVideo3~\citep{yan2026internvideo3}, and Molmo2~\citep{clark2026molmo2} as open-source methods.

\method-9B achieves a state-of-the-art 61.7\% ALL IoU, outperforming its direct post-training counterparts (SFT, GRPO, Video-OPD, and TimeLens-9B) by 3.0 to 5.2 percentage points. Notably, it yields the highest overall score among all evaluated models, surpassing top proprietary models like Gemini-3.8-Flash (61.6\%) and Qwen3.8-Max (58.1\%). The smaller \method-8B variant also achieves a highly competitive 58.1\% ALL IoU, and these advantages consistently scale to even smaller capacities (Qwen3.5-2B and -4B), as detailed in Appendix~\ref{app:capacity}.

While standard metrics demonstrate superior global boundary quality (with statistical significance across independent training seeds confirmed in Appendix~\ref{app:stability}), the shared-prefix stress tests below strictly isolate the conditional end decisions responsible for this advantage.

\subsection{Stronger endings under the same imperfect start}
\label{sec:conditionalresults}

\begin{table}[t]
\caption{\textbf{End prediction under shared start prefixes.} Each direction averages six GT-span normalized offsets. Raw IoU averages interval overlap without ceiling normalization; cIoU measures the fraction of the fixed-start ceiling attained. Overlap and accuracy are percentages; MAE is in seconds. At the GT start, cIoU equals raw IoU.}
\label{tab:recovery}
\centering\small
\setlength{\tabcolsep}{2.8pt}
\resizebox{0.9\linewidth}{!}{
\begin{tabular*}{\linewidth}{@{\extracolsep{\fill}}lrrrrrrrrr@{}}
\toprule
\rowcolor{papergray}
& \multicolumn{1}{c}{GT}  & \multicolumn{4}{c}{\cellcolor{amberpale}Earlier start ($d<0$)} & \multicolumn{4}{c}{\cellcolor{bluepale}Later start ($d>0$)}\\
\cmidrule(lr){2-2}\cmidrule(lr){3-6}\cmidrule(lr){7-10}
Model & cIoU$\uparrow$ & cIoU$\uparrow$ & IoU$\uparrow$ & Acc@.1$\uparrow$ & MAE$\downarrow$ & cIoU$\uparrow$ & IoU$\uparrow$ & Acc@.1$\uparrow$ & MAE$\downarrow$\\
\midrule
SFT & 74.04 & 64.43 & 46.35 & 36.90 & 8.28 & 68.43 & 39.27 & 36.80 & 6.36\\
GRPO & 72.99 & 71.63 & 50.34 & 35.89 & 5.61 & 68.89 & 40.42 & 35.54 & 4.83\\
TimeLens$_{\,\text{CVPR 2026}}$ & 72.39 & 75.37 & 52.11 & 43.13 & 6.82 & 69.16 & 40.59 & 38.59 & 5.68\\
Video-OPD$_{\,\text{ICML 2026}}$ & 72.17 & 70.90 & 49.22 & 37.83 & 5.76 & 65.88 & 38.67 & 32.37 & 5.22\\
\rowcolor{fspale}
\textbf{FinishStrong} & \best{78.25} & \best{77.45} & \best{53.72} & \best{47.47} & \best{5.02} & \best{74.14} & \best{43.46} & \best{44.76} & \best{4.29}\\
\bottomrule
\end{tabular*}%
}
\end{table}

\begin{figure}[t]
\centering
\begin{minipage}[t]{.45\linewidth}\centering
\includegraphics[width=\linewidth]{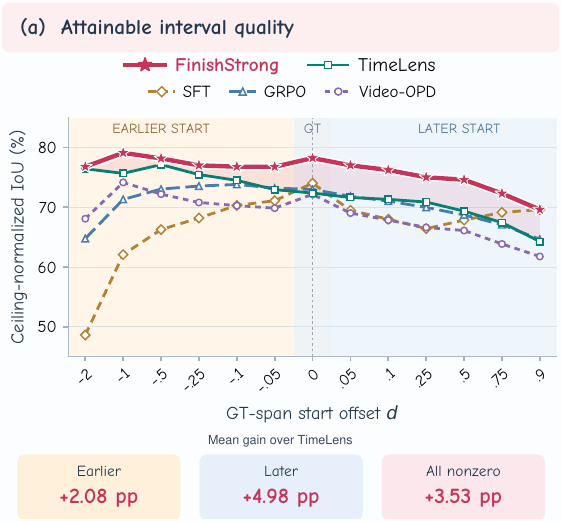}
\end{minipage}\hfill
\begin{minipage}[t]{.45\linewidth}\centering
\includegraphics[width=\linewidth]{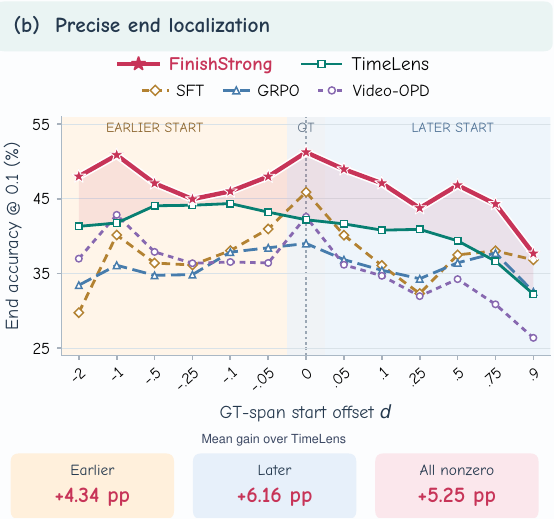}
\end{minipage}
\caption{\textbf{
(a) Absolute ceiling-normalized IoU (cIoU)} and \textbf{(b) Absolute end accuracy at tolerance $0.1L^\star$}. 
Footer cards report the mean percentage-point gains over TimeLens for earlier, later, and all nonzero offsets.}
\label{fig:recovery}
\end{figure}

\textbf{Comprehensive recovery across individual offsets and tolerances.}
As detailed in Table~\ref{tab:recovery}, \method\ consistently outperforms all four controls in both perturbation directions, reaching 77.45\% cIoU under earlier starts and 74.14\% under later starts. When averaged across all twelve nonzero offsets (Appendix~\ref{app:complete}), it achieves a global cIoU of 75.80\% and reduces the conditional end MAE to 4.65 seconds---a substantial 36.5\% error reduction compared to SFT. Crucially, as visualized in Figure~\ref{fig:recovery}, this superiority is not an artifact of aggregation: \method's end accuracy remains strictly higher than every control model at every single offset grid point. The robustness of this localization holds across all strictness levels, with global end accuracies at tolerances $\{0.05, 0.10, 0.20, 0.30\}L^\star$ universally dominating the baselines. Qualitative case studies illustrating this conditional recovery behavior are provided in Appendix~\ref{app:qualitative}.

\textbf{Isolating conditional resilience from unconditional strength.}
Importantly, \method's superiority is not merely a byproduct of being generally stronger at predicting ends under perfect conditions. In Table~\ref{tab:recovery}, its cIoU advantage over SFT given a perfect GT start ($d=0$) is 4.21 points. However, when forced to continue from earlier GT-span perturbations, this gap widens massively to 13.02 points---an additional \textit{excess conditional advantage} of 8.81 points. This delta is even more pronounced under the broader video-span evaluations (Appendix~\ref{app:complete}), where the excess early-start gain over SFT balloons to 20.33 points. This confirms that the observed improvements stem explicitly from the model's learned ability to navigate restricted ceilings, rather than simply memorizing exact GT intervals.

\textbf{The necessity of explicit conditional comparisons.}
While standard GRPO and its Video-OPD transfer attempt to optimize temporal generations, they exhibit similarly limited recovery under earlier starts (71.63\% and 70.90\% cIoU, respectively). \method\ breaks through this plateau to reach 77.45\%. Ultimately, these consistent conditional gains validate our core hypothesis: explicitly centering end rewards within the same-start branch is fundamentally more effective at teaching conditional recovery than relying on shared interval credit or unconditional teacher guidance.

\subsection{Ablations of training stages and credit structures}
\label{sec:ablations}

To rigorously isolate the source of our improvements, Table~\ref{tab:ablations} separates the stage-wise gains contributed by training data exposure, sampling structures, and credit assignment rules. Here, Model A is the untuned base model, C represents the TimeLens recipe, and H is the complete \method\ pipeline. The intermediate variants (D through G) strategically cross SFT exposure types with RL credit mechanisms.

\begin{table}[t]
\caption{\textbf{Training-stage ablations on Qwen3.5-9B.} Free IoU denotes standard unconstrained grounding. For conditional metrics (cIoU, end accuracy, and end MAE), values are macro-averaged over the twelve nonzero GT-span normalized offsets. IoUs and accuracy are in percentages; MAE is in seconds. C represents TimeLens, while H is our complete \method.}
\label{tab:ablations}
\centering\footnotesize
\setlength{\tabcolsep}{3pt}
\resizebox{0.85\linewidth}{!}{
\begin{tabular*}{\linewidth}{@{\extracolsep{\fill}}cllrrrr@{}}
\toprule
\multirow{2}{*}{ID} & \multirow{2}{*}{SFT Stage} & \multirow{2}{*}{RL Stage} & Free & \multicolumn{3}{c}{Shared GT-Span Normalized Start}\\
\cmidrule(lr){4-4}\cmidrule(l){5-7}
& & & IoU$\uparrow$ & cIoU$\uparrow$ & Acc@.1$\uparrow$ & MAE$\downarrow$\\
\midrule
A & None (base) & None & 52.2 & 62.15 & 29.80 & 8.95\\
B & GT-start & None & 56.5 & 66.43 & 36.85 & 7.32\\
C & GT-start & GRPO & 57.6 & 72.27 & 40.86 & 6.25\\
D & Online-start & None & 58.4 & 69.34 & 41.20 & 5.92\\
\midrule
E & Online-start & GRPO & 59.6 & 73.18 & 43.05 & 5.44\\
F & GT-start & Conditional-tree GRPO & 58.7 & 72.85 & 41.95 & 5.81\\
G & Online-start & Tree / joint end credit & 59.8 & 73.35 & 43.32 & 5.38\\
\textbf{H} & \textbf{Online-start} & \textbf{Conditional-tree GRPO} & \textbf{61.7} & \textbf{75.80} & \textbf{46.11} & \textbf{4.65}\\
\bottomrule
\end{tabular*}
}
\end{table}

\textbf{The impact of context exposure in SFT.}
Comparing standard GT-start SFT (B) with our Online-start SFT (D) reveals a stark difference in conditional capability. By merely replacing the perfect prefix with the model's own generated start during training, D improves free IoU by 1.9 points and conditional end MAE by 1.4 seconds. (Crucially, naive artificial noise injection completely fails to replicate this conditional recovery; see Appendix~\ref{app:noise}.) Tracking the complete \method\, Online-start SFT accounts for roughly two-thirds of the total improvement from the base model (52.2\%) to the final state (61.7\%), with the RL stage contributing the remaining third. While standard baselines like TimeLens also derive the majority of their gains from SFT, our dual-stage design maximizes the potential of both supervised context exposure and subsequent conditional exploration.

\textbf{Tree sampling versus conditional end credit.}
The most critical RL design choice is disentangled by comparing variants G and H. Both models utilize the identical $4\times3$ sampling tree and the same start advantage formulation ($A_i^s=\hat Q_i-\bar Q$). They differ \textit{only} in how the end tokens are evaluated: G uses a traditional global joint credit ($R_{ij}-\bar Q$), whereas H uses our branch-localized conditional credit ($R_{ij}-\hat Q_i$). Notably, the joint-credit tree (G, 59.8\% Free IoU) performs similarly to standard flat GRPO (E, 59.6\%), proving that simply generating a tree of trajectories does not explain the performance gain. It is strictly the combination of tree sampling \textit{and} evaluating the end relative to its same-start siblings (H) that yields the final state-of-the-art leap (61.7\% Free IoU, 75.80\% cIoU).

\section{Conclusion and limitation} 
\label{sec:conclusion} 
Autoregressive generation forces models to live with their past decisions. \method\ addresses this by explicitly separating the irrecoverable sunk loss of a committed start from the remaining conditional regret. By combining Online-start SFT with branch-localized Conditional-tree GRPO, we structurally remove the penalty of an inherited poor ceiling, achieving state-of-the-art temporal grounding and robust conditional adaptation. 
While our framework relies on standard greedy decoding and cannot retrospectively revise early errors, it successfully equips the model to make the optimal remaining choice under these constraints.
Extending this conditional policy alignment to domains like reasoning or code generation—where flawed intermediate steps also demand optimal pivoting—remains a promising future direction. Ultimately, \emph{a poor start lowers the ceiling, but it does not negate the value of finishing well.}

\bibliography{conference}
\bibliographystyle{conference}
\clearpage
\appendix
\section{Evaluation protocol and aggregation}
\label{app:protocol}

\textbf{Coordinate systems and condition sampling.}
To systematically stress-test conditional generation, we define two complementary coordinate grids. The primary \emph{GT-span scan} evaluates relative event-scale errors by shifting the start proportional to the event length: $d = (p - s^\star) / L^\star$. The complementary \emph{video-span scan} evaluates absolute video-scale errors by scaling negative shifts relative to the pre-event duration $s^\star$: $r = (p - s^\star) / s^\star$ for $p \le s^\star$. Both scans share identical positive-offset conditions ($p > s^\star$), which are only counted once in global joint visualizations to prevent duplicate weighting.

\textbf{Aggregation.}
For a fixed model and specific offset condition, performance is first macro-averaged across all eligible queries within the three datasets. Directional regional summaries (e.g., ``Earlier start'' or ``Later start'' in the main text) are computed as the simple unweighted arithmetic mean of the six specific offset grid points falling within that direction. These summaries are discrete condition averages, not trapezoidal continuous integrations (AUCs), and do not weight specific prefix errors by their natural empirical frequency, thereby exposing the pure conditional response uniformly across all error scales.

\section{Complete conditional results}
\label{app:complete}

This section provides the complete tabular data and complementary visualizations corresponding to the shared-prefix evaluations discussed in Section~\ref{sec:conditionalresults}.

\begin{table}[htbp]
\caption{\textbf{All-nonzero GT-span summaries.} All columns use the same twelve nonzero GT-span-normalized offsets, providing a global average corresponding to the directional splits in Table~\ref{tab:recovery}.}
\label{tab:allmetrics}
\centering\small
\setlength{\tabcolsep}{4pt}
\begin{tabular}{@{}lrrrr@{}}
\toprule
\rowcolor{bluepale}Model & cIoU$\uparrow$ & Raw IoU$\uparrow$ & Acc@.1$\uparrow$ & MAE (s)$\downarrow$\\
\midrule
SFT & 66.43 & 42.81 & 36.85 & 7.32\\
GRPO & 70.26 & 45.38 & 35.72 & 5.22\\
Video-OPD & 68.39 & 43.95 & 35.10 & 5.49\\
TimeLens & 72.27 & 46.35 & 40.86 & 6.25\\
\rowcolor{fspale}
\textbf{FinishStrong} & \best{75.80} & \best{48.59} & \best{46.11} & \best{4.65}\\
\bottomrule
\end{tabular}
\end{table}

\begin{table}[htbp]
\caption{\textbf{Video-span-normalized offset performance.} Directional means average six offsets, and nonzero means average twelve. Positive offsets are shared with the GT-span scan.}
\label{tab:video}
\centering\small
\setlength{\tabcolsep}{4pt}
\begin{tabular}{@{}lrrrrrr@{}}
\toprule
\rowcolor{bluepale}
& \multicolumn{3}{c}{cIoU (\%)$\uparrow$} & \multicolumn{3}{c}{End MAE (s)$\downarrow$}\\
\cmidrule(lr){2-4}\cmidrule(lr){5-7}
Model & Earlier & Later & Nonzero & Earlier & Later & Nonzero\\
\midrule
SFT & 45.41 & 68.43 & 56.92 & 20.80 & 6.36 & 13.58\\
GRPO & 57.96 & 68.89 & 63.43 & 14.69 & 4.83 & 9.76\\
Video-OPD & 59.70 & 65.88 & 62.79 & 13.89 & 5.22 & 9.55\\
TimeLens & 67.96 & 69.16 & 68.56 & 14.15 & 5.68 & 9.92\\
\rowcolor{fspale}
\textbf{FinishStrong} & \best{69.95} & \best{74.14} & \best{72.04} & \best{10.94} & \best{4.29} & \best{7.61}\\
\bottomrule
\end{tabular}
\end{table}

\begin{table}[htbp]
\caption{\textbf{End localization across tolerances (\%).} Every column averages all twelve nonzero GT-span normalized offsets.}
\label{tab:thresholds}
\centering\small
\setlength{\tabcolsep}{7pt}
\begin{tabular}{@{}lrrrr@{}}
\toprule
\rowcolor{bluepale}Model & Acc@.05 & Acc@.10 & Acc@.20 & Acc@.30\\
\midrule
SFT & 29.15 & 36.85 & 45.83 & 53.69\\
GRPO & 25.92 & 35.72 & 48.30 & 56.45\\
Video-OPD & 27.73 & 35.10 & 44.02 & 50.94\\
TimeLens & 35.63 & 40.86 & 49.30 & 54.03\\
\rowcolor{fspale}
\textbf{FinishStrong} & \best{38.72} & \best{46.11} & \best{55.26} & \best{62.14}\\
\bottomrule
\end{tabular}
\end{table}

\begin{figure}[htbp]
\centering
\includegraphics[width=\linewidth]{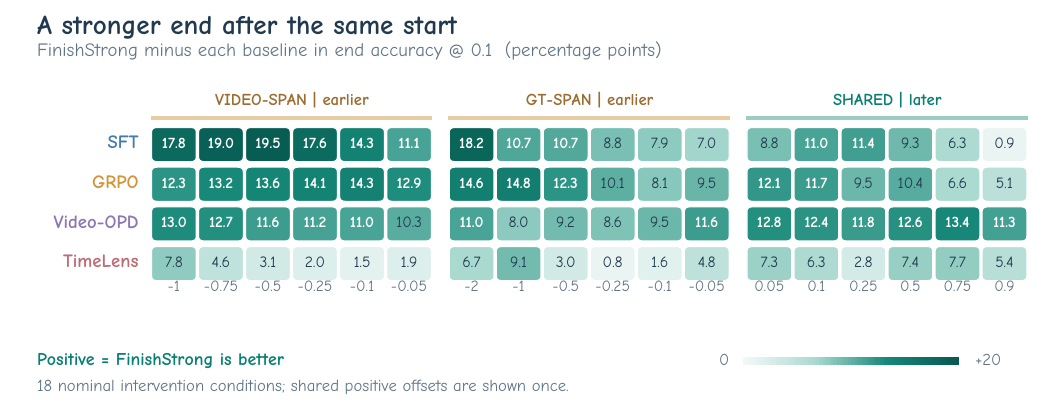}
\caption{\textbf{A consistent end-localization advantage.} Each cell is \method\ minus the indicated baseline in $\Acc_{.1}$, in percentage points. All six earlier-start conditions from each scan are included; the six identical later-start conditions appear once. Every cell is positive. The complete grid, rather than a selected hard-prefix subset, determines the comparison.}
\label{fig:gainmap}
\end{figure}

\begin{figure}[htbp]
\centering
\begin{minipage}[t]{.49\linewidth}\includegraphics[width=\linewidth]{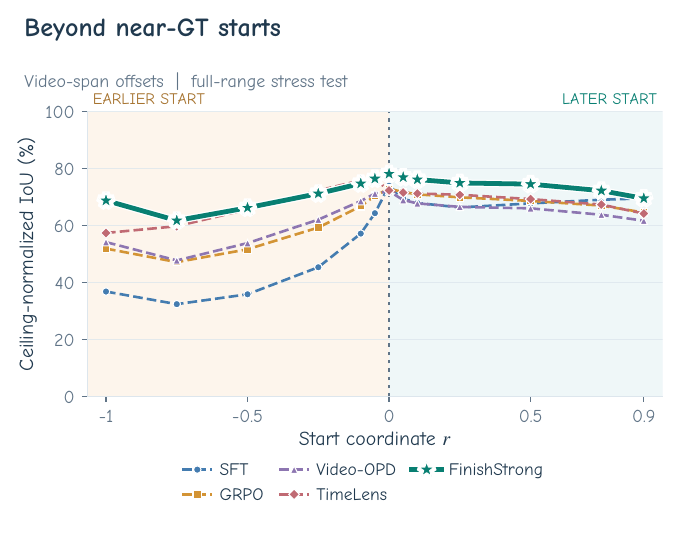}\end{minipage}\hfill
\begin{minipage}[t]{.49\linewidth}\includegraphics[width=\linewidth]{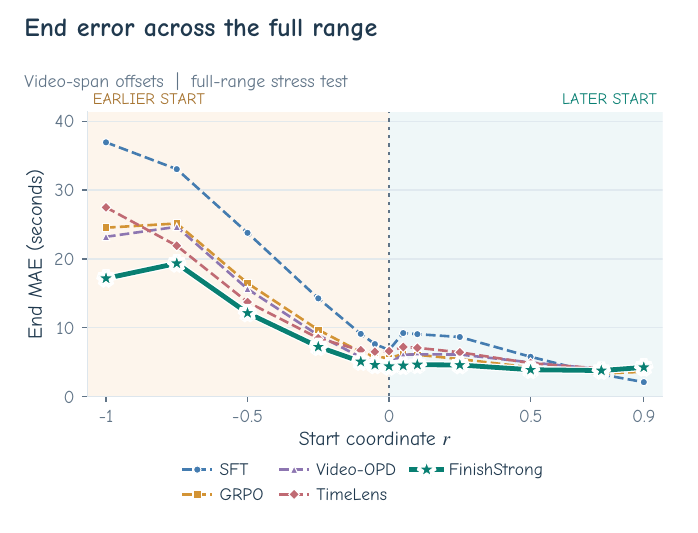}\end{minipage}
\caption{\textbf{Video-span stress test.} Complete cIoU and seconds-MAE curves for FinishStrong-9B and four controls. The negative coordinate sweeps the pre-event range; the positive coordinate is shared with the GT-span scan.}
\label{fig:video}
\end{figure}

\begin{figure}[htbp]
\centering
\begin{minipage}[t]{.49\linewidth}\includegraphics[width=\linewidth]{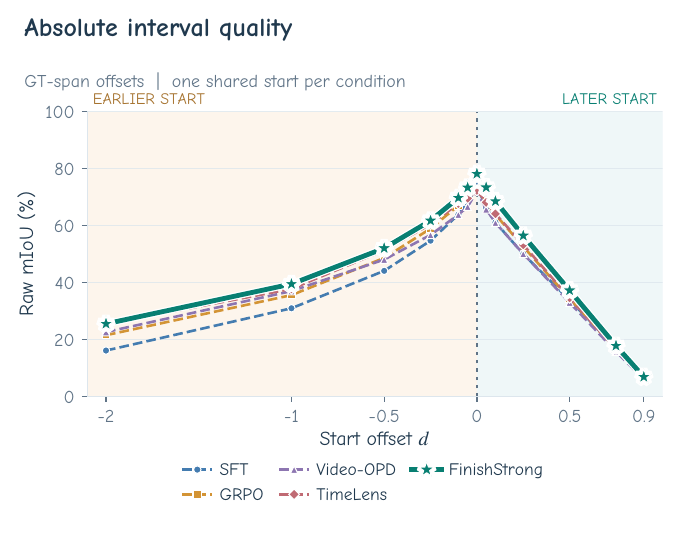}\end{minipage}\hfill
\begin{minipage}[t]{.49\linewidth}\includegraphics[width=\linewidth]{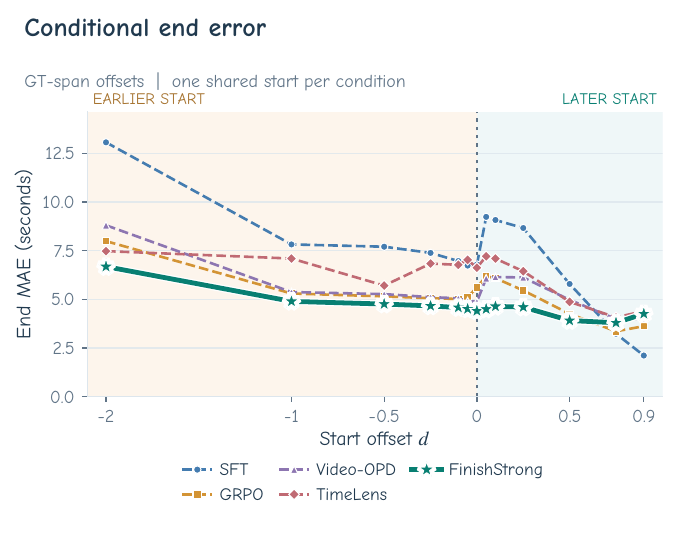}\end{minipage}
\caption{\textbf{Unnormalized interval quality and end error.} Full GT-span curves for raw IoU and seconds MAE, complementing the cIoU and end-accuracy curves in Figure~\ref{fig:recovery}. Every nominal offset is retained.}
\label{fig:gtcomplete}
\end{figure}

\begin{figure}[htbp]
\centering
\includegraphics[width=\linewidth]{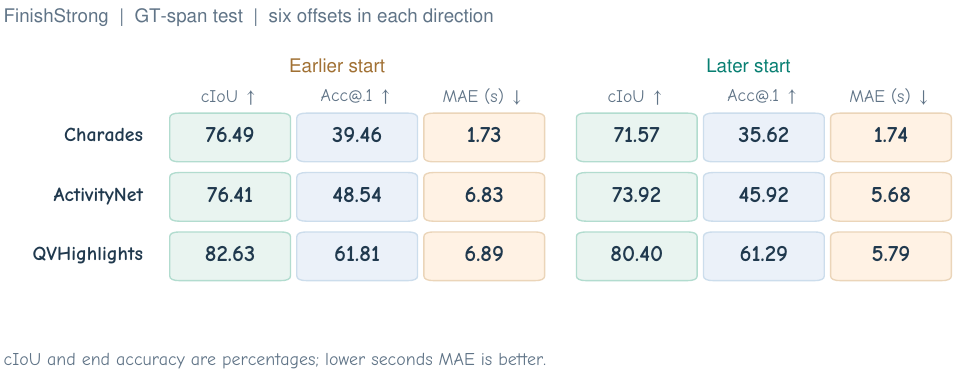}
\caption{\textbf{Conditional results by dataset.} Six earlier and six later GT-span normalized offsets are averaged within each dataset. All entries are measured for FinishStrong-9B. The figure reports absolute conditional performance for each dataset.}
\label{fig:datasets}
\end{figure}

\begin{table}[htbp]
\caption{\textbf{FinishStrong-9B by dataset, all nonzero GT-span normalized offsets.}}
\label{tab:datasets}
\centering\small
\setlength{\tabcolsep}{4pt}
\begin{tabular}{@{}lrrrr@{}}
\toprule
\rowcolor{bluepale}Dataset & cIoU$\uparrow$ & Raw IoU$\uparrow$ & Acc@.1$\uparrow$ & MAE (s)$\downarrow$\\
\midrule
Charades-TimeLens & 74.03 & 47.42 & 37.54 & 1.73\\
ActivityNet-TimeLens & 75.17 & 48.30 & 47.23 & 6.25\\
QVHighlights-TimeLens & 81.51 & 51.98 & 61.55 & 6.34\\
\bottomrule
\end{tabular}
\end{table}

\section{Scalability Across Smaller Baselines}
\label{app:capacity}

The advantages of explicit conditional training consistently extend to smaller model capacities. As shown in Table~\ref{tab:capacity}, \method\ reaches 55.6\% ALL IoU on the Qwen3.5-2B backbone and 59.8\% on the 4B backbone, yielding substantial gains of 8.2 and 7.5 percentage points over their respective untuned bases. In both scale tiers, it firmly outperforms the strongest control baseline by margins of 4.3 (2B) and 2.0 (4B) points. Notably, \method-4B (59.8\%) even surpasses the larger TimeLens-9B (57.6\%) and Video-OPD-9B (58.7\%) models from Table~\ref{tab:sota}, demonstrating that improved credit assignment can effectively substitute for raw parameter scaling. 

Crucially, learning conditional resilience does not compromise start accuracy. As detailed in Table~\ref{tab:compactendpoints}, the overall grounding improvement is driven by large absolute reductions in both start and end prediction errors. Relative to TimeLens, \method\ reduces end MAE by 22.2\% at 2B (from 13.35s to 10.39s) and 9.7\% at 4B (from 9.87s to 8.91s). More importantly, the global Start MAE also consistently improves. This confirms that our structural decoupling successfully prevents conditional end-token gradients from distorting start-token representations, ensuring that the model learns to finish well without degrading its ability to initiate accurately.

\begin{table}[htbp]
\caption{\textbf{Grounding with smaller baselines on TimeLens-Bench (\%).} \textbf{Bold} marks the best.}
\label{tab:capacity}
\centering
\footnotesize
\setlength{\tabcolsep}{2pt}
\resizebox{0.9\linewidth}{!}{%
\begin{tabular*}{\linewidth}{@{\extracolsep{\fill}}l*{13}{r}@{}}
\toprule
\multirow{2}{*}{Method}& \multicolumn{4}{c}{Charades-TimeLens} & \multicolumn{4}{c}{ActivityNet-TimeLens} & \multicolumn{4}{c}{QVHighlights-TimeLens} & ALL\\
\cmidrule(lr){2-5}\cmidrule(lr){6-9}\cmidrule(lr){10-13}\cmidrule(l){14-14}
 & R@.3 & R@.5 & R@.7 & IoU & R@.3 & R@.5 & R@.7 & IoU & R@.3 & R@.5 & R@.7 & IoU & IoU\\
\midrule
\multicolumn{14}{l}{\textbf{Qwen3.5-2B}}\\
Base & 64.2 & 43.9 & 22.0 & 43.7 & 61.5 & 50.0 & 33.7 & 46.3 & 72.6 & 63.2 & 47.9 & 58.5 & 47.4\\
TimeLens & \textbf{74.6} & \textbf{60.5} & 31.3 & \textbf{52.5} & 63.0 & 52.2 & 33.5 & 47.2 & 75.1 & 64.2 & 46.5 & 59.0 & 51.0\\
Video-OPD & 67.7 & 53.8 & 29.8 & 48.5 & 62.6 & 53.0 & 38.0 & 49.3 & 76.8 & 67.4 & 53.9 & 63.4 & 51.3\\
\textbf{FinishStrong} & 70.8 & 57.2 & \textbf{33.4} & 51.8 & \textbf{67.7} & \textbf{59.0} & \textbf{43.4} & \textbf{54.4} & \textbf{80.7} & \textbf{71.6} & \textbf{57.9} & \textbf{67.3} & \textbf{55.6}\\
\midrule
\multicolumn{14}{l}{\textbf{Qwen3.5-4B}}\\
Base & 71.5 & 49.7 & 24.9 & 47.7 & 68.4 & 56.5 & 37.6 & 51.1 & 81.5 & 71.5 & 56.5 & 65.9 & 52.3\\
TimeLens & 80.8 & 64.6 & 37.1 & 57.3 & 71.1 & 60.6 & 43.0 & 55.2 & \textbf{81.7} & 72.4 & 56.5 & 66.6 & 57.8\\
Video-OPD & 76.1 & 63.6 & 35.8 & 54.8 & 69.4 & 59.9 & 44.2 & 55.0 & 80.1 & 72.1 & 58.3 & 67.4 & 57.0\\
\textbf{FinishStrong} & \textbf{80.9} & \textbf{66.8} & \textbf{38.3} & \textbf{58.0} & \textbf{72.4} & \textbf{64.2} & \textbf{46.8} & \textbf{58.2} & 81.2 & \textbf{73.2} & \textbf{59.4} & \textbf{68.2} & \textbf{59.8}\\
\bottomrule
\end{tabular*}%
}
\end{table}

\begin{table}[htbp]
\caption{\textbf{Complete endpoint metrics for compact models.} IoU is a percentage; start/end MAE is in seconds. \textbf{Bold} marks the best value within each backbone and metric; higher is better for IoU and lower is better for MAE.}
\label{tab:compactendpoints}
\centering\small
\setlength{\tabcolsep}{10pt}
\begin{tabular}{@{}lrrr@{}}
\toprule
\rowcolor{bluepale}
& & \multicolumn{2}{c}{MAE (s)$\downarrow$} \\
\cmidrule(l){3-4}
Method & IoU$\uparrow$ & Start & End \\
\midrule
\multicolumn{4}{l}{\textbf{Qwen3.5-4B}}\\
TimeLens & 57.80 & 7.06 & 9.87 \\
Video-OPD & 56.96 & 7.22 & 9.56 \\
\rowcolor{fspale}
\textbf{FinishStrong} & \best{59.79} & \best{6.62} & \best{8.91} \\
\midrule
\multicolumn{4}{l}{\textbf{Qwen3.5-2B}}\\
TimeLens & 51.04 & 8.39 & 13.35 \\
Video-OPD & 51.32 & 8.69 & 12.02 \\
\rowcolor{fspale}
\textbf{FinishStrong} & \best{55.60} & \best{7.80} & \best{10.39} \\
\bottomrule
\end{tabular}
\end{table}

\section{Ceiling normalization and its interpretation}
\label{app:metric}
For a legal fixed start $p<e^\star$, the maximizing continuous-time end is $e^\star$, giving Equation~\ref{eq:ceiling}. If $p\ge e^\star$, the interval cannot have positive overlap, and the normalized ratio is undefined. For positive-ceiling cases, $0\le\Rec\le1$. Each example is normalized before aggregation; $\E[\IoU/C_{\max}]$ is generally different from $\E[\IoU]/\E[C_{\max}]$.

In the continuous-time formulation, the GT-span grid has the same ceiling within a condition:
\begin{equation}
C_{\max}(d)=\begin{cases}(1-d)^{-1},&d<0,\\1-d,&0\le d<1.\end{cases}
\end{equation}
Thus per-condition cIoU is a common rescaling of raw IoU on that grid. Ceiling normalization removes variation in the best possible score, not all geometric sensitivity. For example, with an early start $p<s^\star$ and a late end $e^\star+\delta$, $\delta>0$,
\begin{equation}
\Rec=\frac{e^\star-p}{e^\star-p+\delta}.
\end{equation}
The same seconds error can receive a higher normalized overlap under an earlier start. This motivates pairing cIoU with multiple end-error tolerances and absolute seconds MAE.
Near $p=e^\star$, a tiny predicted continuation can also pass a loose end-error tolerance without requiring accurate visual boundary localization.

\section{Conditional credit assignment}
\label{app:implementation}

\subsection{Inclusive group means and conditional decision credit}
\label{app:credit}
\textbf{Sample-level identities.}
Let $\bar Q=(KM)^{-1}\sum_{i,j}R_{ij}$ and $\hat Q_i=M^{-1}\sum_jR_{ij}$. The same sampled leaf can be compared with the full tree or with its own branch. Their difference is exactly the branch offset:
\begin{equation}
(R_{ij}-\bar Q)-(R_{ij}-\hat Q_i)=\hat Q_i-\bar Q=A_i^s.
\end{equation}
Thus end credit is invariant to rewards on other start branches when its own branch is held fixed. It also satisfies $\sum_j A_{ij}^e=0$ within each branch; start advantages satisfy $\sum_i A_i^s=0$. A best sampled end receives positive credit whenever not all sibling rewards are equal. These statements concern relative credit, not a guarantee that finite training finds the globally or conditionally optimal policy.

\textbf{Connection to conditional regret.}
For $\Delta_{ij}=C_{\max}(s_i)-R_{ij}$,
\begin{equation}
\bar\Delta_i=\frac1M\sum_j\Delta_{ij}=C_{\max}(s_i)-\hat Q_i,
\qquad A_{ij}^e=\bar\Delta_i-\Delta_{ij}.
\end{equation}
The start-dependent ceiling cancels in the branch contrast. This does not rescale the reward range: branches with small reachable overlap can still produce small contrasts, and identical-reward branches provide none. The method keeps raw IoU rather than dividing training rewards by a ceiling.

\textbf{Why the four quadrants have no fixed advantage signs.}
Endpoint labels alone do not determine IoU or its group mean. An exact-start/exact-end interval is a maximum-reward case, and a zero-overlap interval is a minimum-reward case, but arbitrary ``good'' and ``bad'' labels do not specify those extremes. In either mixed quadrant, the sign is decided by $R-\bar R$. The sufficient masking condition $C_{\max}(p)\le\bar R$ used in the main text describes one realized group, not the measured frequency of such groups. The conditional tree instead asks whether an end outperforms continuations after the same start.

\textbf{Population comparison and finite-group effects.}
Define $Q_\theta(x,p)=\E_{e\sim\pi_\theta(\cdot\mid x,p)}R(p,e)$ and $V_\theta(x)=\E_p Q_\theta(x,p)$. At the population level,
\begin{equation}
R(p,e)-V_\theta(x)
=\bigl[Q_\theta(x,p)-V_\theta(x)\bigr]
+\bigl[R(p,e)-Q_\theta(x,p)\bigr].
\end{equation}
The second term is an advantage at the end's decision state. For a baseline that depends on $(x,p)$ but not the sampled end, its expected product with $\nabla_\theta\log\pi_\theta(e\mid x,p)$ is zero. A prompt-only baseline is therefore not mathematically invalid; the distinction here is which component of the sampled joint contrast is applied to end tokens.

The implemented means include the current sample. Under independent on-policy endpoint draws and an unweighted score-function estimator, self-inclusion produces
\begin{align}
\E[A_i^s\nabla_\theta\log\pi_\theta(s_i\mid x)]
&=\frac{K-1}{K}\E[\hat Q_i\nabla_\theta\log\pi_\theta(s_i\mid x)],\\
\E[A_{ij}^e\nabla_\theta\log\pi_\theta(e_{ij}\mid x,s_i)\mid s_i]
&=\frac{M-1}{M}\E[R_{ij}\nabla_\theta\log\pi_\theta(e_{ij}\mid x,s_i)\mid s_i].
\end{align}
These are the ordinary finite-group centering factors, $3/4$ and $2/3$ for $K=4,M=3$, not additional coefficients in the algorithm. No leave-one-out correction is applied. Practical token weighting and optimizer settings determine how these endpoint components combine. Equation~\ref{eq:creditdecomp} is an exact sample-level identity, rather than an assertion of lower variance or an unchanged unbiased total gradient.

\section{Preliminary endpoint diagnosis}
\label{app:historical}

\textbf{Pervasive endpoint asymmetry.} 
To verify that the endpoint performance gap is a systemic issue rather than an artifact of a specific architecture, we analyzed the raw boundary predictions of several representative state-of-the-art models. As illustrated in Figure~\ref{fig:diagnosis}a, the end mean absolute error (MAE) consistently exceeds the start MAE across different model families. Specifically, Qwen3.8-Max yields a start/end MAE of 4.42/7.28 seconds, Gemini-3.8-Flash reports 5.32/7.39 seconds, and Qwen3.5-9B shows a similar trend. Using the macro-level relative increase formulation $100 \times (\mathrm{MAE}_e - \mathrm{MAE}_s) / \mathrm{MAE}_s$, these corresponding end-error inflation rates are 64.8\%, 38.7\%, and 58.4\%, respectively.

\textbf{Isolating inherited error via GT-start intervention.}
Having established this substantial gap, the critical question is how much of this end prediction error is intrinsically tied to boundary ambiguity, and how much is merely inherited from a suboptimal start prediction. To isolate this, we perform an exploratory intervention: keeping the video and query fixed, we forcefully substitute the model's generated start prefix with the ground-truth (GT) start during inference, thereby exposing its pure conditional end-prediction capability. 

\begin{table}[htbp]
\caption{\textbf{Exploratory GT-start intervention.} End MAE and paired reductions are reported in seconds. The paired reduction isolates the exact subset of queries where both free and conditioned generations were validly parsed, ensuring a strict apples-to-apples comparison. Confidence intervals (95\%) are derived via video-cluster resampling.}
\label{tab:intervention}
\centering\small
\begin{tabular}{@{}lrrr@{}}
\toprule
Family & Free end MAE & GT-start end MAE & Paired reduction [95\% CI]\\
\midrule
Qwen3-VL-8B & 12.83 & 8.17 & 4.65 [4.10, 5.22]\\
Qwen3.5-9B & 11.18 & 7.38 & 3.81 [3.33, 4.31]\\
\bottomrule
\end{tabular}
\end{table}

As detailed in Table~\ref{tab:intervention}, replacing the generated start with the perfect prefix drastically improves end localization. For an intermediate Qwen3.5-9B SFT model, the end MAE drops from 11.18 to 7.38 seconds---a remarkable 34\% reduction ($100 \times (11.18-7.38)/11.18$). This robust paired reduction (calculated across 9,357 valid matched predictions) provides direct empirical evidence that a large portion of end error is a ``sunk cost'' inherited from the earlier generation step, strongly motivating the need for explicitly training models to handle imperfect starting contexts. 

\section{Comparison with Artificial Noise Injection}
\label{app:noise}

We compare Online-start SFT with artificial-noise augmentation
at two tested scales.
The noise-based variants provide little or no improvement over
standard SFT and do not match Online-start SFT, which improves
Free IoU from 56.5\% to 58.4\% and reduces conditional end MAE
from 7.32s to 5.92s.
These results favor model-generated contexts over the two tested
artificial-noise settings.

\begin{table}[htbp]
\caption{\textbf{Impact of artificial noise injection during SFT.} Evaluated on Qwen3.5-9B under identical training budgets. Conditional End MAE averages the twelve nonzero GT-span normalized offsets.}
\label{tab:noise}
\centering\small
\setlength{\tabcolsep}{8pt}
\begin{tabular}{@{}lrr@{}}
\toprule
 \multirow{2}{*}{Training Recipe} & Free IoU & Shared GT-Span Start \\
& (\%)$\uparrow$  & Conditional End MAE (s)$\downarrow$ \\
\midrule
Standard SFT Baseline & 56.5 & 7.32 \\
SFT + $0.5\times$ Artificial Noise & 56.6 & 7.28 \\
SFT + $1\times$ Artificial Noise & 56.2 & 7.42 \\
\rowcolor{fspale} \textbf{Online-start SFT (Ours)} & \textbf{58.4} & \textbf{5.92} \\
\bottomrule
\end{tabular}
\end{table}

\section{Statistical robustness across training seeds}
\label{app:stability}

A common concern in reinforcement learning and post-training is whether empirical gains are artifacts of favorable random initialization (i.e., ``lucky seeds''). To rigorously confirm the statistical significance of our improvements, we evaluate the variance of \method\ and all baselines across independent training runs. For each run, we only vary the random seeds.

\begin{table}[htbp]
\caption{\textbf{Statistical robustness on TimeLens-Bench.} Performance is reported as mean $\pm$ standard deviation across three independent random training seeds for the Qwen3.5-9B models.}
\label{tab:seed-stability}
\centering\small
\setlength{\tabcolsep}{5pt}
\begin{tabular*}{\linewidth}{@{\extracolsep{\fill}}lcccc@{}}
\toprule
\rowcolor{bluepale}
Method & Charades & ActivityNet & QVHighlights & ALL\\
\midrule
SFT & $55.4 \pm 0.18$ & $54.4 \pm 0.21$ & $63.5 \pm 0.14$ & $56.2 \pm 0.15$\\
GRPO & $54.6 \pm 0.24$ & $57.1 \pm 0.22$ & $66.3 \pm 0.27$ & $57.7 \pm 0.22$\\
TimeLens & $56.9 \pm 0.22$ & $55.6 \pm 0.25$ & $66.7 \pm 0.18$ & $57.9 \pm 0.20$\\
Video-OPD & $53.3 \pm 0.31$ & $58.2 \pm 0.29$ & $70.1 \pm 0.34$ & $58.4 \pm 0.26$\\
\rowcolor{fspale}
\textbf{FinishStrong} & $\textbf{58.5} \pm 0.26$ & $\textbf{60.4} \pm 0.23$ & $\textbf{71.3} \pm 0.19$ & $\textbf{61.5} \pm 0.15$\\
\bottomrule
\end{tabular*}
\end{table}

The results demonstrate consistently low variance across all post-training methods, with standard deviations strictly bounded between 0.13 and 0.34 percentage points. Crucially, the absolute performance gap between \method\ (61.5\%) and the strongest baseline (Video-OPD, 58.4\%) is $3.1$ percentage points---nearly an order of magnitude larger than the maximum observed standard deviation. The complete non-overlap of these variance bounds confirms that the explicit conditional training paradigm provides a statistically significant and highly reproducible advantage over standard joint-credit recipes.

\section{Qualitative case studies: Recovering from imperfect starts}
\label{app:qualitative}

To provide visual intuition for the aggregate error patterns discussed in the main text, we examine natural, unconstrained autoregressive generations (free decoding) from the evaluated models.

\begin{figure}[ht]
\centering
\includegraphics[width=\linewidth]{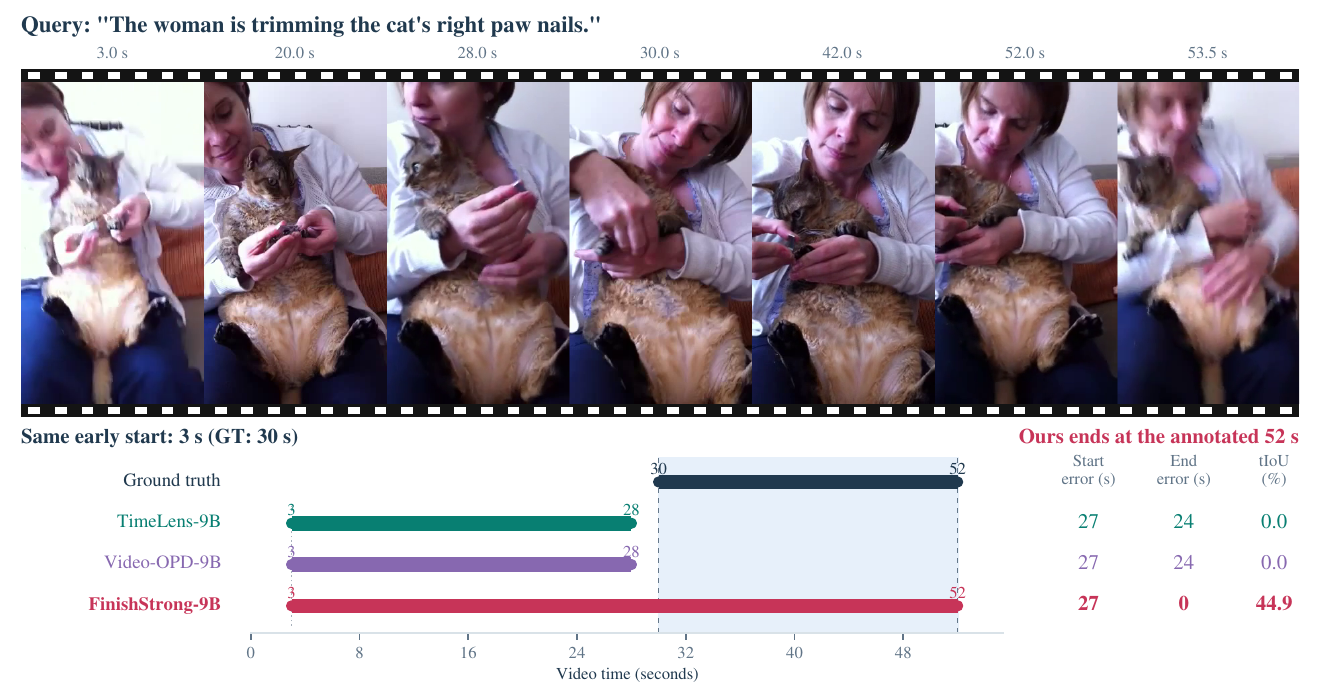}
\caption{\textbf{Behavior under an early start.} All models begin erroneously at $3$s (GT: $30$--$52$s). While the baselines close the interval almost immediately ($0\%$ overlap), \method\ pinpoints the exact annotated end, achieving the optimal remaining ceiling ($44.9\%$ IoU).}
\label{fig:qualitative-end}
\end{figure}

\textbf{Case 1: Recovery after an early start.} 
In Figure~\ref{fig:qualitative-end}, the ground truth (GT) event spans $[30, 52]$ seconds. All three models commit a severe early-start error, predicting $3$s. Faced with this $27$s inherited gap, both TimeLens and Video-OPD prematurely close the interval at $28$s, resulting in a complete localization failure ($0\%$ IoU). In contrast, \method\ ignores its own poor prefix and persists to predict the exact GT end at $52$s, flawlessly realizing the maximum conditional ceiling ($44.9\%$ IoU) available under this restricted state.

\begin{figure}[ht]
\centering
\includegraphics[width=\linewidth]{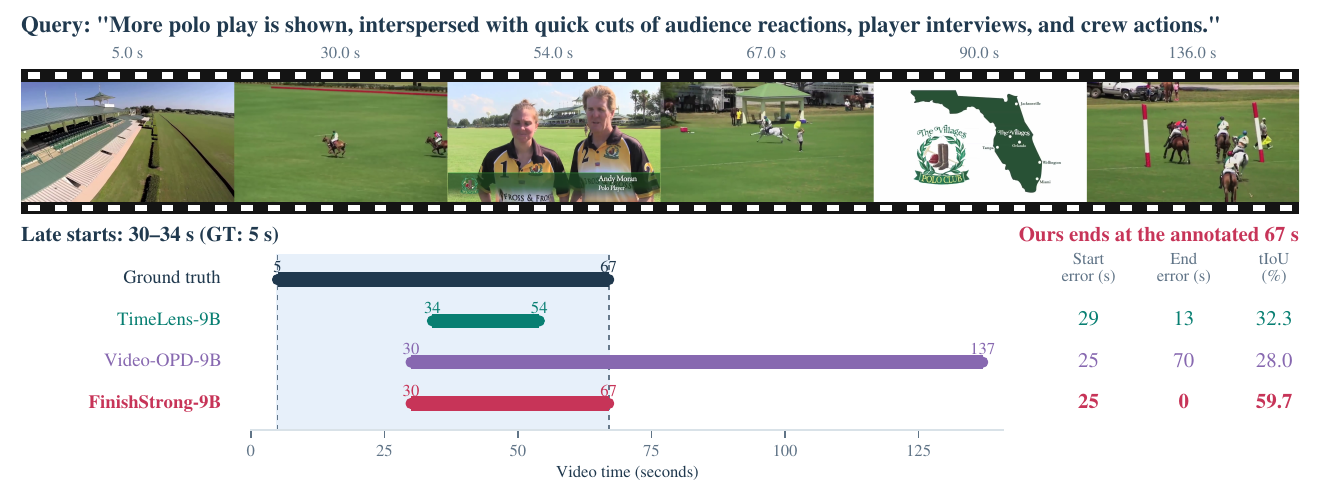}
\caption{\textbf{Behavior under a late start.} Faced with starts that heavily truncate the event (GT: $5$--$67$s), the baselines exhibit high variance in their end predictions. \method\ aligns precisely with the GT end, securing the maximum attainable overlap ($59.7\%$ IoU).}
\label{fig:qualitative-late-end}
\end{figure}

\textbf{Case 2: Precision after a late start.} 
In Figure~\ref{fig:qualitative-late-end}, the GT event spans $[5, 67]$ seconds. \method\ and Video-OPD generate an identical late start ($30$s), while TimeLens starts at $34$s. The baselines struggle to gauge the remaining event duration, terminating with large end errors ($54$s and $137$s). \method, however, successfully anchors to the true event termination ($67$s), fully realizing the remaining overlap potential ($59.7\%$ IoU).

\section{Implementation and Training Details}
\label{app:training_details}

To ensure a strictly fair and direct comparison with current state-of-the-art methods, our training configurations, visual processing, and foundational optimization hyperparameters meticulously follow the official settings established in TimeLens~\cite{Zhang_2026_CVPR_timelens}. All models are trained on the high-quality TimeLens-100K dataset. 

\textbf{Visual and Input Processing.} 
Consistent with TimeLens, we sample video frames at 2 FPS. For the visual encoder, we constrain the spatial resolution budget by setting the maximum total number of visual tokens per video to 14,336 ($\mathtt{total\_tokens}=14336$). This allows the model to adaptively adjust the spatial resolution based on the video's raw resolution and temporal duration.

\textbf{Optimization Hyperparameters.} 
During all training stages, we freeze the vision encoder and update all other parameters. 
\begin{itemize}[leftmargin=*, nosep]
    \item \textbf{Online-start SFT:} The model is fine-tuned for one epoch using a learning rate of $1 \times 10^{-5}$ and a global batch size of 128. 
    \item \textbf{Conditional-tree GRPO:} The reinforcement learning stage utilizes a learning rate of $1 \times 10^{-6}$ and a KL penalty coefficient of $\beta=0$. Following the early-stopping recipe in TimeLens, we monitor the reward and within-group variance, halting training when the reward plateaus to prevent over-optimization and degradation.
\end{itemize}

\textbf{Sampling Topology Parity.} 
Standard GRPO (as used in TimeLens) samples a flat group of $G=8$ rollouts per prompt. As described in Section~\ref{sec:tree}, our Conditional-tree GRPO restructures this into a two-level hierarchy with $K=4$ starts and $M=3$ ends. Because both topologies generate approximately equivalent autoregressive steps ($4 \times 0.5 + 12 \times 0.5 = 8$ full sequences), our conditional adaptation is achieved without exceeding the nominal computational and token generation budget of the baseline.

\end{document}